\documentclass[runningheads]{llncs}
\usepackage{graphicx}
\usepackage{enumerate}
\usepackage{cite}
\usepackage{amsmath}
\usepackage{float}   
\usepackage{multirow}
\usepackage{hyperref}

\begin{document}
\title{Approaches for Improving the Performance of Fake News Detection in Bangla: Imbalance Handling and Model Stacking}
\titlerunning{Approaches for Imbalance Handling and Model Stacking}

\author{Md. Muzakker Hossain\thanks{The authors contributed equally to this paper.} \and
Zahin Awosaf$^\star$ \and
Md. Salman Hossan Prottoy \and\\ Abu Saleh Muhammod Alvy \and Md. Kishor Morol}

\authorrunning{Hossain et al.}

\institute{American International University-Bangladesh, Dhaka, Bangladesh.\\
\{18-37801-2, 18-37064-1, 18-36902-1, 18-37487-1\}@student.aiub.edu, \\and kishor@aiub.edu}
\maketitle
\begin{abstract}
Imbalanced datasets can lead to biasness into the detection of fake news. In this work, we present several strategies for resolving the imbalance issue for fake news detection in Bangla with a comparative assessment of proposed methodologies. Additionally, we propose a technique for improving performance even when the dataset is imbalanced. We applied our proposed approaches to BanFakeNews \cite{hossain2020banfakenews}, a dataset developed for the purpose of detecting fake news in Bangla comprising of 50K instances but is significantly skewed, with 97\% of majority instances. We obtained a 93.1\% F1-score using data manipulation manipulation techniques such as SMOTE, and a 79.1\% F1-score using without data manipulation approaches such as Stacked Generalization. Without implementing these techniques, the F1-score would have been 67.6\% for baseline models. We see this work as an important step towards paving the way of fake news detection in Bangla. By implementing these strategies the obstacles of imbalanced dataset can be removed and improvement in the performance can be achieved.

\end{abstract}
\section{Introduction}
The dissemination of misinformation, colloquially referred to as Fake News, is escalating as more people gain access to the internet. These information can be fed to us in the form of news, stories, or social media posts. These false information can be categorized into many forms such as, Clickbait, Propaganda, Satire/Parody, Biased/Slanted News, and so on. These fabricated news can have a range of effects, from political polarization to communal violence. A study shows that people tend to share the falsified news more than the authentic ones in the mainstream social media \cite{vosoughi2018spread}. Automated programs used to engage in social media or Social Bots that impersonate human behavior, are more frequently used in popular social media. Separate reports have shown that between 9 to 15\% of active twitter accounts are Social Bots and about 60 million bots were found in Facebook \cite{varol2017online}. And these automated accounts can be used to manipulate other users of these social platforms and spread misinformation. The impacts of falsified information are spread all over the world. In Bangladesh many unfortunate incidents like communal violence occur because of some rumors, or false information shared on mainstream social media. Hence, it has become critical to identify these fake news.\\
\indent
A big hindrance of fake news detection can be improper or lack of datasets. Compared to prominent languages not many significant datasets can be found for Bangla. The only comprehensive dataset available in Bangla \cite{hossain2020banfakenews}, has a big hindrance of data imbalance, meaning one class is significantly larger than the other. As a result, this is a significant setback in the route of research for detecting fake news in Bangla which can hinder the objective of tackling the risk of falsified information and fight against rumors.\\
\indent
So in this work, we focus to remove the obstacle and pave the way for research in detecting fake news in Bangla.\\
\textbf{Contributions.} The bridge the current gap we-
\begin{enumerate}[•]
  \item Propose several methods to handle the data imbalance
  \item Make a comparative study between the proposed models to find out the best suited method for data imbalance handling
  \item Show an alternative way to optimize the performance in detecting fake news without any data manipulation, namely Stacked Generalization
\end{enumerate}
We strongly hope to contribute adequate information to overcome the limitations of the BanFakeNews dataset and make a significant contribution to the Bangla fake news detection system.

\section{Related Work}
The lack of a substantial dataset is the primary impediment to developing an automated system for detecting fake news in Bangla. Possessing a proper dataset can pave the way for potential research in this field. This issue is not apparent in popular languages such as English, which is why numerous substantial studies have been undertaken in this domain for English.\\
\indent 
For example, \cite{nakamura2019r} presents a dataset containing over 1 million samples of fake news in English. Another prominent dataset is \cite{wang2017liar} which is a news dataset for the purpose of investigating automated fake news detection using surface-level linguistic patterns. \\
\indent
Scarcity of datasets is one of the reasons why not so many researches in fake news detection in Bangla were conducted. To our best knowledge \cite{hossain2020banfakenews} is the only proper dataset available for fake news detection purpose in Bangla. Although the number of authentic news is approximately 37.47 times higher than the number of fake news, it offers a significant amount of data for a language with scarce resources like Bangla. Since it is a comparatively new dataset in Bangla, not many researches have been conducted based on this dataset. The experimental analysis of \cite{sraboni2021fakedetect}'s proposed model on real-world data demonstrates that the Passive Aggressive Classifier and Support Vector Machine achieve 93.8\% and 93.5\% accuracy, respectively. And this research was conducted based on \cite{hossain2020banfakenews} dataset.\\
\indent
\cite{padurariu2019dealing} used a wide range of resampling schemes to deal with the imbalanced dataset, including random oversampling, SMOTE, SMOTE-SVM, WESMOTE, and CSMOTE. In \cite{chawla2002smote}, a state-of-the-art oversampling technique called the Synthetic Minority Over-sampling Technique (SMOTE) was implemented. \cite{he2008adasyn} implements the Adaptive Synthetic (ADASYN) oversampling method, which generates new synthetic instances based on difficult-to-learn instances of the minority class. An oversampling method to deal with imbalanced data was implemented in \cite{nguyen2011borderline} which mainly concentrates on the borderline instances that are crucial for establishing decision boundary.\\
\indent
Another efficient technique for handling imbalanced dataset can be model stacking. While a single classifier is inefficient, stacking multiple generalizers can increase a model's performance by reducing the generalization error rate of one or more generalizers \cite{wolpert1992stacked}. This approach is termed Stacked Generalization, as presented by \cite{wolpert1992stacked}. The researchers demonstrated that by stacking models, the performance of each model can be significantly improved \cite{jiang2021novel}.

\section{Background}
In this section, we discuss the necessary background information related to our work.

\subsection{Dataset}
BanFakeNews \cite{hossain2020banfakenews}, a publicly available annotated dataset consists of $\approx$50K Bangla news [Table: \ref{tab:banfake}]. Owing to the lack of a well-established dataset for Bangla fake news, we used BanFakeNews to perform text classification in Bangla with the aim of detecting fake news. The dataset collected the authentic news from 22 most popular and mainstream trusted news portals in Bangladesh \cite{hossain2020banfakenews}. The fake news were commonly attributed for all kinds of satirical, clickbait, and falsified information.
\begin{table}
\setlength{\tabcolsep}{7pt} 
\renewcommand{\arraystretch}{1} 
\begin{center}
\caption{Features of BanFakeNews Dataset}\label{tab1}
\label{tab:banfake}
\begin{tabular}{|c|c|c|}
\hline
\textbf{Class Name} &  \textbf{Total Rows} & \textbf{Percentage of Total}\\
\hline
Authentic & 48,678 & 97.4\%\\
\hline
Fake & 1,299 & $\approx$2.6\%\\
\hline
\end{tabular}
\end{center}
\end{table}
As the [Table: \ref{tab:banfake}] indicates, this dataset is utterly imbalanced containing only 2.6\% of fake news. Therefore, in order to minimize this disparity, we utilized several strategies and compared the outcome to the baseline model. It's worth noting that the dataset comprises a variety of features, including the source, the source's domain, the headline, the date the article was published, and the article's category. However, we have used the article's text as our primary feature.

\section{Methodology}
In this section, we describe the data preprocessing, feature extraction from raw texts of dataset, and proposed techniques to handle the imbalance issue.

\subsection{Data Preprocessing}
\label{Sec:datapreprocess}
As we are dealing with text classification the dataset had to be prepared in a certain way before feeding to the models. Data preprocessing is a strategy for transforming unstructured data into a usable and effective format. Initially the text contained many unnecessary emoticons, symbols, punctuation, digits, and foreign alphabets. So, we preprocessed the dataset by removing the unnecessary fragments at first, followed by stop words removal and stemming.\\
\textbf{Standardization:}
The unnecessary fragments such as, punctuation marks, symbols (e.g., \@, \#, \$, \%, \&), English alphabets, Bangla and English digits, and emoticons have been omitted since they are irrelevant to text classification.\\
\textbf{Stopwords Removal:}
Stop words are frequent but insignificant, these words do not deliver any necessary information and they make no contribution to the text classification process. So to prune off these words, we used a stop words library called "Stopwords Bengali (BN)" from github \footnote{\url{https://github.com/stopwords-iso/stopwords-bn}}, which is a complete set of Bengali stopwords to make the process efficient. Some common stopwords are shown below:
\begin{figure}[H]
\centering
\includegraphics[width=15em, height=0.6cm]{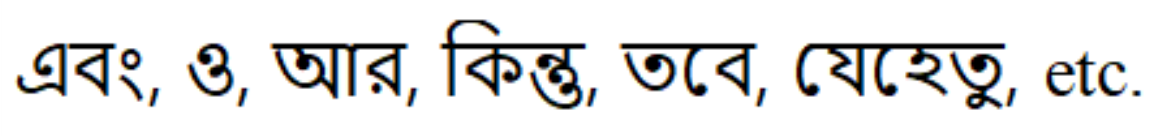}
  \label{fig:stopwords}
\end{figure}

\textbf{Stemming:}
Stemming is the process of finding the root of a word. This makes the process even more efficient. We have used the python Bangla Stemmer\footnote{\url{https://pypi.org/project/bangla-stemmer/}} package to do the stemming.

\subsection{Feature Extraction}
Algorithms for machine learning are limited to numerical values. To implement machine learning over text, we should first transform our text data as vector values which is known as feature extraction or vectorization. We have used two of the most widely used feature extraction techniques for text classification– Count Vectorizer and TF-IDF Vectorizer since these are not language dependent and can simply be incorporated into machine learning models.\\ \\
\indent
\textbf{Count Vectorizer:}
Count Vectorizer creates a sparse matrix with columns for unique words and rows for text samples. The value of each cell corresponds to the number of occurring words in the text data.\\ \\
\indent
\textbf{TF-IDF Vectorizer:}
TF-IDF or Term Frequency–Inverse Document Frequency method computes a word's distinctiveness by comparing the number of times a word appears in a document to the total number of documents in which the word appears. Count Vectorizer counts the occurrences of a certain word in a document, whereas the TF-IDF also considers the number of documents in which the word appears. 
For our work, we have used n-gram range(1, 2) which is a continuous sequence of n items extracted from a particular text or speech sample.

\subsection{Structure of Models}
In our work, initially, we developed the baseline model in order to establish a benchmark for performance measurement. Then three strategies namely Oversampling, Under-sampling, and Modifying the Class-Weight; were applied to handle the imbalanced data. Finally, an ensemble technique “Stacked Generalization” was further performed to improve the outcome of the models.\\ \\
\noindent
\textbf{Baseline Model}\\
Six classifiers (i.e., Logistic Regression, Support Vector Machine, Multinomial Naïve Bayes, Bernoulli Naïve Bayes, Random Forest Classifier, and Decision Tree Classifier) were chosen as baseline models in order to create a benchmark for further model testing.\\ \\
\noindent
\textbf{Oversampling}\\
As shown in [Table: \ref{tab:banfake}], the dataset included only 2.6 percent of the minority class. To compensate for the remaining 97.4 percent, the first technique we utilized was oversampling. Oversampling is a technique for resolving data imbalances that increases the sample size of minority class. We’ve performed three kinds of oversampling methods namely, Random Oversampling, SMOTE, and ADASYN.\\
\indent
\begin{figure}[h]
\centering
\includegraphics[width=\textwidth]{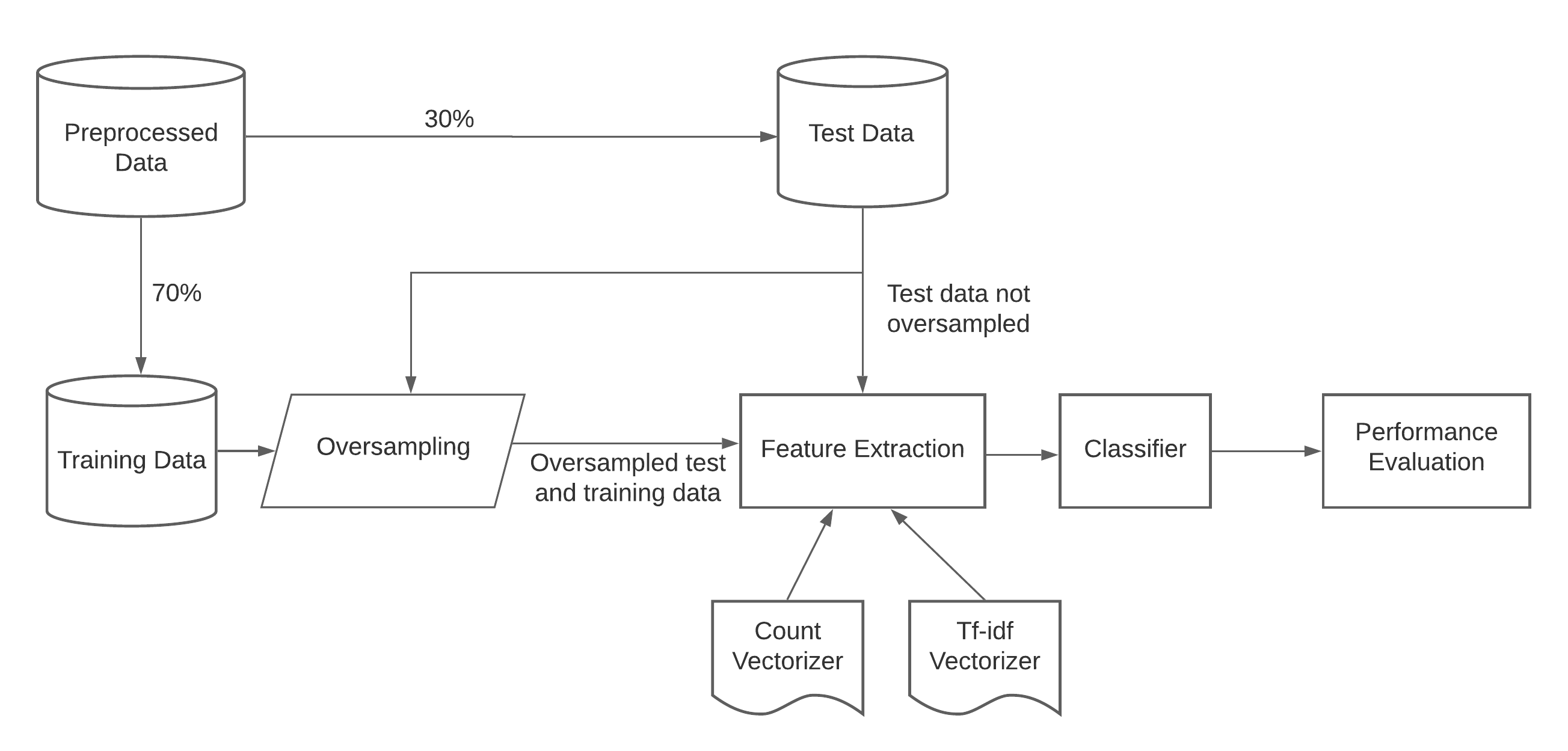}
  \caption{Illustration of Oversampling Technique}
  \label{fig:oversampling}
  \centering
\end{figure}

\textbf{Random Oversampling:} Random Oversampling is a naive approach for generating new samples by randomly sampling current data samples with substitution.\\ \\
\indent
\textbf{SMOTE:} Synthetic Minority Oversampling Technique (SMOTE) is an oversampling method that generates synthetic samples from the minority class. SMOTE begins by selecting random data from the minority class, and then generates synthetic data using a k-nearest neighbor method.\\ \\
\indent
\textbf{ADASYN:} Adaptive Synthetic Sampling Method or simply ADASYN, is a technique for adaptively producing minority data samples based on their distributions: more synthetic data is produced for minority class samples that are more difficult to learn than for minority class samples that are simpler to learn.
\begin{table}[H]
\setlength{\tabcolsep}{7pt} 
\renewcommand{\arraystretch}{1} 
\begin{center}
\caption{Oversampled Data}
\label{tab:oversampled}
\begin{tabular}{{|c|c|c|}}
\hline
 & \textbf {Before 
Oversampling
}& \textbf{After 
Oversampling
}\\
\hline
      Train Data (Authentic)& 34,075& 34,075 \\
\hline
      Train Data (Fake)& 909& 34,075 \\
\hline
      Test Data (Authentic)& 14,603& 14,603 \\
\hline
      Test Data (Fake)& 390& 14,603 \\      
\hline
\end{tabular}
\end{center}
\end{table}
Prior to oversampling [Figure: \ref{fig:oversampling}], data for each technique were separated in a 70:30 ratio as train:test data to eliminate the chance of data contamination. Without doing so, the same data used for training and testing can be mixed together, resulting in a biased output. The overview of oversampled data is shown on [Table: \ref{tab:oversampled}].\\
\noindent
\textbf{Undersampling}\\
Under-sampling is the reciprocal of over-sampling, in which instances from the majority class are discarded in order to reduce the disparity between the majority and minority classes. We performed two different types of undersampling: Random Undersampling and NearMiss.\\ \\
\indent
\textbf{Random Undersampling:} Random undersampling is the process of randomly choosing and discarding samples from the majority class from the training dataset.\\ \\
\indent
\textbf{NearMiss:} This technique examines the distribution of the larger class and randomly discards samples from it. When two points in the distribution belong to different classes and are extremely close to each other, this technique eliminates the data points from the larger class, attempting to balance the distribution.\\ \\
\indent
The total preprocessed data were first undersampled and then were split into 80:20 ratio for train and test data accordingly. 
By following undersampling, the majority class was reduced to 1,299 rows, same as the minority class, and the data were split 80:20.\\ \\
\textbf{Modifying Class-Weight}\\
The aim of modifying class weights is to penalize the minority class for misclassification by increasing class weights while decreasing weights for the majority class. In order to calculate class-weight, we have used balanced as the argument for class-weight function.\\ \\
\noindent
\textbf{Model Stacking}\\
\begin{figure}[H]
\centering
\includegraphics[width=\textwidth]{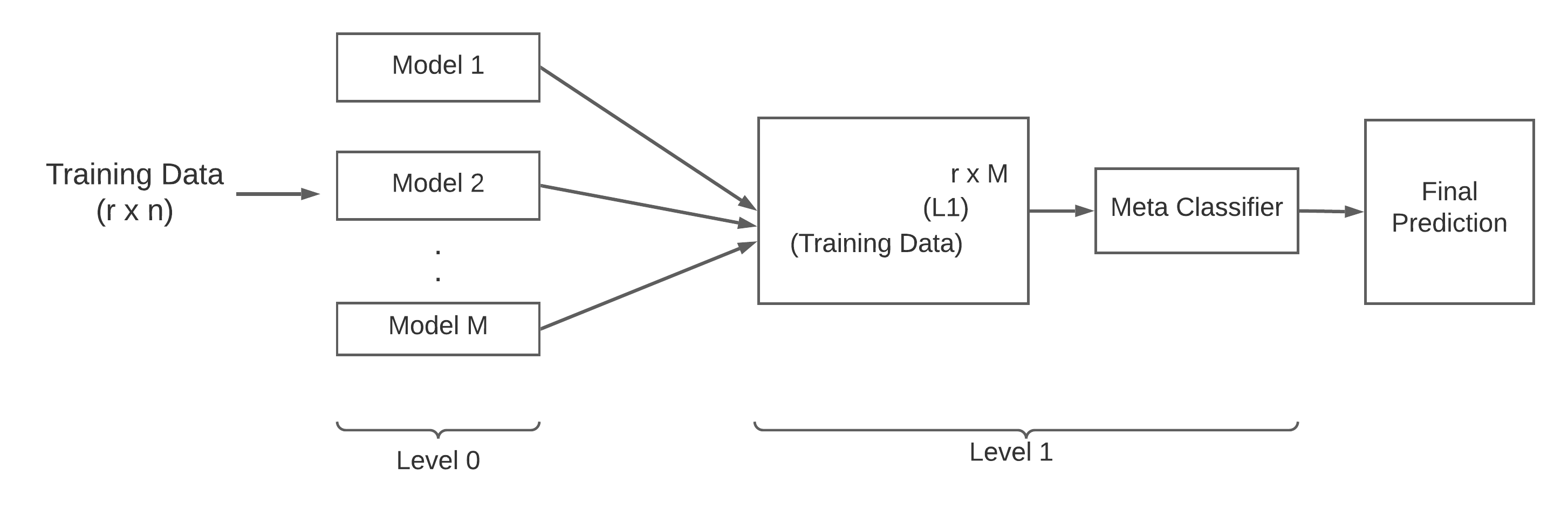}
  \caption{Mathematical Demonstration of Model Stacking}
  \label{fig:stacking}
  \centering
\end{figure}
Apart from resolving imbalances, another strategy that may be used to increase the model's performance is Model Stacking. Model stacking is a strategy of combining several machine learning models to have a better output. Model stacking can be completed by different ensemble techniques. In this work, we have implemented one of the most renowned ensemble technique known as Stacked Generalization, or stacking which employs a new model to discover the optimal way to combine predictions from a set of models trained on the same dataset. We executed the strategy by building a set of six models for level-0 and then picking one of them as the meta-classifier for level-1 [Figure: \ref{fig:stacking}]. This process was repeated six times.
Initially, the training data contained r observations, and n features. The different models were cross-validated (cv = 5) and each model produced outcome that was then cast into meta-classifier, meaning the M predictions became features (r x M) in the second level (Level-1). The second level model or meta-classifier were then trained again and tested for final prediction.

\subsection{Experimental Setup}
To begin with, we divide the preprocessed data [Section: \ref{Sec:datapreprocess}] into train and test sets using an 80:20 train-to-test split for all techniques except Oversampling which was a split of 70:30. Then, using the Count Vectorizer and the TF-IDF vectorizer, both the train and test data were transformed into vector forms. Following that, all six classifiers were fit to the train data and then evaluated against the test data.\\
\indent
The parameter list is as follows, for Logistic Regression the parameters are Solver: Saga, Multi-Class: Multinomial; for Support Vector Classification Kernel: rbf, Gamma: 1; for both of the Naïve Bayes method Alpha: 0.01; for Random Forest Classifier N-Estimators: 400, Max-Feature: sqrt; and for Decision Tree Classifier Criterion: Gini, Max-Depth: 6. Finally, in stacking: we used 5-folds for cross validation generator.

\section{Results \& Discussion}
In this section, we describe in-depth performance analysis of our proposed models. We calculate four performance indicators to evaluate the models’ efficiency: Accuracy, Precision, Recall, and F1-score. Since the dataset was significantly imbalanced, we mainly focused on F1-score, which tend to better reflect the scenario of minority classes in an imbalanced dataset.

\subsection{Overall Performance of All Methods}
\begin{table}[h]
\setlength{\tabcolsep}{6pt} 
\renewcommand{\arraystretch}{1} 
    \centering
        \caption{Overall Performance of All Methods}
        \label{tab:allmodel}
        \begin{tabular}{|c|c|c|c|}
        \hline
         \textbf{Method}& \textbf{Feature Extraction}& \textbf{Classifier}& \textbf{F1-Score} \\ 
         \hline
        \multirow{2}{*}{\textbf{Baseline}}& Count& LR& 0.604 \\
        \cline{2-4}
        & TF-IDF& LR& 0.676\\
        \cline{1-4}
        \multirow{2}{*}{\textbf{Random Oversampling (N)}}& Count& LR& 0.452 \\
        \cline{2-4}
        & TF-IDF& LR& 0.703\\
        \cline{1-4}
        \multirow{2}{*}{\textbf{Random Oversampling}}& Count& MNB& 0.845 \\
        \cline{2-4}
        & TF-IDF& LR& 0.826\\
        \cline{1-4}
        \multirow{2}{*}{\textbf{SMOTE (N)}}& Count& MNB& 0.476 \\
        \cline{2-4}
        & TF-IDF& LR& 0.696\\
        \cline{1-4}
        \multirow{2}{*}{\textbf{SMOTE}}& Count& LR& \textbf{0.931} \\
        \cline{2-4}
        & TF-IDF& LR& 0.863\\
        \cline{1-4}
        \multirow{2}{*}{\textbf{ADASYN (N)}}& Count& MNB& 0.460 \\
        \cline{2-4}
        & TF-IDF& LR& 0.695\\
        \cline{1-4}
        \multirow{2}{*}{\textbf{ADASYN}}& Count& BNB& 0.914 \\
        \cline{2-4}
        & TF-IDF& LR& 0.866\\
        \cline{1-4}
        \multirow{2}{*}{\textbf{Random Undersampling}}& Count& BNB, RFC& 0.893 \\
        \cline{2-4}
        & TF-IDF& MNB& 0.911\\
        \cline{1-4}
        \multirow{2}{*}{\textbf{Near-Miss}}& Count& RFC& \textbf{0.943} \\
        \cline{2-4}
        & TF-IDF& SVM& 0.935\\
        \cline{1-4}
        \multirow{2}{*}{\textbf{Modifying Class-Weight}}& Count& LR& 0.493 \\
        \cline{2-4}
        & TF-IDF& LR& 0.757\\
        \cline{1-4}
        \textbf{Model Stacking}& TF-IDF& RFC& 0.791\\
        \hline
        \end{tabular}
\end{table}

Here [Table: \ref{tab:allmodel}], (N) indicates that the test data for that specific method were not oversampled, only the test data obtained by 70:30 (train:test) split was used. The column "Classifier" denotes the classifier that produced the highest F1-score for the specified method.\\
\indent
A baseline model is a straightforward model that produces decent outcome without any changes in the algorithm. As a result, it can serve as a benchmark for other models constructed using the same dataset. For the performance matrix, we focus on the F1-score because accuracy isn't a viable statistic for an imbalanced dataset \cite{galar2011review}. And as for Precision and Recall, F1-score is the harmonic mean of both of these metrics and so F1-score is our ultimate statistic for evaluation. In our work we’ve emphasized on minority class while computing F1-Score. As there is vast disparity between majority and minority classes and so the majority class has nearly perfect F1-score.\\
\indent
As the [Table: \ref{tab:allmodel}] indicates, while undersampling methods (i.e., random undersampling and near-miss) generate better outcomes, we do not consider them to be the optimum method for dealing with data imbalance. Since undersampling is the process of excluding data from the majority class, this might result in the omission of important instances from the dataset, hence influencing the desired outcome. Additionally, when previously unseen or new data are presented, undersampling models perform poorly. \\
\indent
Due to the random replication of the data in the case of random oversampling, the generated outcome may be biased, as the same instances are copied repeatedly and the prediction count is repeated for the same instances. However, this is not the case with SMOTE and ADASYN. SMOTE and ADASYN generate new synthetic data, and as shown in the table, when test data is oversampled with SMOTE and ADASYN, the results are significantly more accurate and consistent. \\
\indent
It is seen from the [Table: \ref{tab:allmodel}] that the modifying class-weight is not appropriate for our dataset. It performed worse than the baseline models in the case of Count Vectorizer. The disparity between the two feature extraction strategies is especially evident in the case of class-weight modification, which is typically not apparent in other models. \\ 
\indent
Several studies have been conducted on fake news detection in Bangla but there’s a scarcity of dataset. BanFakeNews \cite{hossain2020banfakenews} was published on 2020 and so not many works have been done utilizing this dataset. \cite{hossain2020banfakenews} proposed a benchmark for this dataset by using neural network models like CNN, LSTM, and BERT. For these models they achieved F1-Score of 0.59, 0.53 and 0.68 respectively for the minority classes whereas we’ve achieved 0.931, 0.914, and 0.791 for methods like SMOTE, ADASYN, and Model Stacking respectively. Other than \cite{hossain2020banfakenews}, to our best knowledge only \cite{sraboni2021fakedetect} had worked on this dataset before us. \cite{sraboni2021fakedetect} had achieved an accuracy of 93.8\% using Passive Aggressive Classifier on this dataset. To overcome the imbalance issue, \cite{sraboni2021fakedetect} worked with 3.5K authentic class instances and manually expanded the fake class instances by merging with another dataset, increasing the class instances to 2.3K. So no proper correlation can be drawn to \cite{sraboni2021fakedetect} with our work.\\
\indent
The other works that were conducted on Bangla fake news detection like \cite{hussain2020detection} had achieved a 96.64\% of accuracy for SVM on their own dataset consisted of around 2541 instances with real and fake being 60.92\% and 39.08\% respectively. In other work, \cite{islam2020bengali} had achieved an accuracy of 85\% for Random Forest Classifier working on their own dataset consisting of 726 news articles. A common phenomena is seen in these studies that their dataset is comparatively much smaller. Comparing to these studies we’ve achieved the highest accuracy of 98.7\% using Logistic Regression and our other models also outperform these studies.

\subsection{Performance of Model Stacking}
Data manipulation can always be a risky manoeuvre whether we apply oversampling or undersampling technique. Unlike prior strategies for resolving imbalances, model stacking does not need to manipulate the dataset. It ensembles the models and enhances the newly created model's performance. The particular stacking technique that we've employed is called Stacked Generalization. As it has been proven that the TF-IDF Vectorizer performs better than the Count Vectorizer as a feature extraction technique, we used the TF-IDF Vectorizer as our feature extraction technique for stacked generalization. We’ve used each of our six classifiers as meta-learner in the process. The result is shown on [Table: \ref{tab:perstack}].
\begin{table}[h]
\setlength{\tabcolsep}{6pt} 
\renewcommand{\arraystretch}{1} 
\caption{Performance of Model Stacking}
    \label{tab:perstack}
    \centering
        \begin{tabular}{|c|c|c|c|c|c|c|c|c|}
        \hline
         & \multicolumn{4}{|c|}{\textbf{TF-IDF Vectorizer}} \\ 
         \hline
        & Accuracy& Precision& Recall& F1-Score \\
        \hline
        LR& 0.988& 0.844& 0.690& 0.759 \\
        \hline
        SVM& 0.989& 0.839& 0.706& 0.767 \\
        \hline
        MNB& 0.981& 0.602& 0.669& 0.675 \\
        \hline
        BNB& 0.978& \textbf{0.923}& 0.142& 0.247 \\
        \hline
        RFC& \textbf{0.990}& 0.846& \textbf{0.742}& \textbf{0.791} \\
        \hline
        DTC& 0.987& 0.776& 0.718& 0.746 \\
        \hline
        \end{tabular}
\end{table}
The outcome of model stacking is much better than the baseline models with a F1-Score of 0.791 for Random Forest as the meta-classifier. Though the outcome is lower than the SMOTE or ADASYN, this method can be utilized for better comprehension and reliability of unseen data. The drawback of this technique is time and space complexity. As several models are stacked upon each other and the dataset is considerably large, it consumed a lot of memory and took much longer time to fit the model and execute the outcome.
\begin{figure}[h]
\centering
\includegraphics[width=10cm, height=4.5cm]{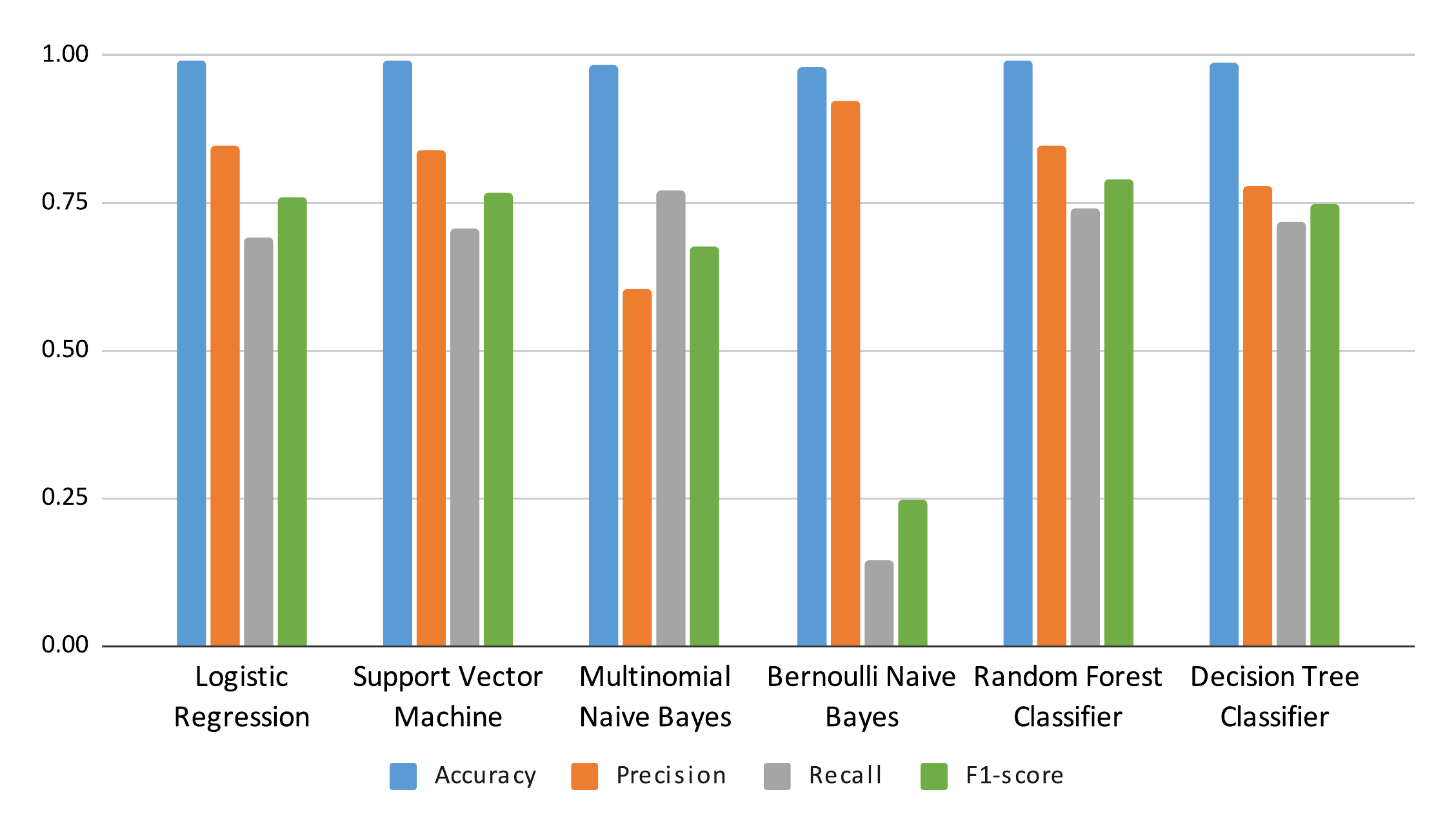}
  \caption{Outcome of Model Stacking}
  \label{fig:perstacking}
  \centering
\end{figure}

\section{Conclusion \& Future Scope}
Fake news detection in Bangla is still in its infancy. In previous studies related to this domain we’ve seen researchers lack a proper dataset to accomplish their objective. Although we found \cite{hossain2020banfakenews} as our primary dataset that contains a sufficient amount of instances, the majority and minority classes are vastly imbalanced. Because of imbalance the expected result can be biased towards the majority class. Thus we focused on addressing the issue and proposed numerous solutions to overcome it. To our best knowledge we are the first to work on the full dataset of BanFakeNews \cite{hossain2020banfakenews}. We’ve not only addressed the imbalance issue of the dataset but also proposed numerous techniques to handle the disparity. Our proposed models successfully surpassed the existing works conducted in Bangla for the purpose of fake news detection in terms of performance metrics and has shown a comparative study between the models. We’ve shown that imbalance handling methods have their own pros and cons. Oversampling methods like SMOTE and ADASYN can be a good solution to reduce the discrepancy between majority and minority classes and the performance of the models also increase with comparison to baseline models [Table: \ref{tab:allmodel}]. Model stacking can be an alternative to data manipulation. A better result can be achieved by using either of these models. Our work aimed to pave the way of expanding research on fake news detection in Bangla. We hope that future studies conducted on this dataset for the purpose of fake news detection, researchers can reliably use any of our proposed imbalance handling techniques.

It's worth noting that we focused exclusively on the text's content and omitted any other features. This is mostly because we have concentrated our efforts on resolving the data imbalance issue. Thus, in the near future, we'd like to work on other dataset features such as source, category, and headline. Additionally, we intend to experiment with different models, including deep learning models such as RNN, CNN, LSTM, and GRU.

\bibliographystyle{splncs04}
\bibliography{327}

\end{document}